\documentclass[letterpaper, 10 pt, conference]{ieeeconf}  

\IEEEoverridecommandlockouts                              

\usepackage{graphicx}
\usepackage{pythonhighlight}
\usepackage{lipsum} 
\usepackage[]{mdframed}
\usepackage{xcolor}
\usepackage{amsmath}
\usepackage{graphicx}
\usepackage{footnote}
\usepackage{amsmath} 
\usepackage{bm}      
\usepackage[ruled,vlined, linesnumbered]{algorithm2e}
\usepackage{amsmath, amssymb}
\usepackage{amsfonts}
\usepackage{algpseudocode}
\usepackage{booktabs}
\usepackage{multirow}
\usepackage{makecell}
\usepackage{romannum}
\usepackage{hyperref}
\usepackage{subcaption}
\DeclareFontFamily{OT1}{pzc}{}
\DeclareFontShape{OT1}{pzc}{m}{it}{<-> s * [1.10] pzcmi7t}{}
\DeclareMathAlphabet{\mathpzc}{OT1}{pzc}{m}{it}

\graphicspath{{./figs/}}

\title{\LARGE \bf
ManiDP: Manipulability-Aware Diffusion Policy for Posture-Dependent Bimanual Manipulation}

\author{Zhuo Li$^{1}$, Junjia Liu$^{1}$, Dianxi Li$^{1}$, Tao Teng$^{1,2}$, Miao Li$^{3}$, Sylvain Calinon$^{4}$, \\ Darwin Caldwell$^{5}$,~\IEEEmembership{Fellow,~IEEE} and Fei Chen$^{*1,2}$,~\IEEEmembership{Senior Member,~IEEE}
\thanks{This work is supported in part by the Research Grants Council of the
Government of the Hong Kong SAR via the Grant 24209021, 14222722, 14213324, 14211723, C7100-22GF and in part by the InnoHK initiative of the Innovation and Technology Commission of the Hong Kong Special Administrative Region Government via the Hong Kong Centre for Logistics Robotics. \textit{(*Corresponding author: Fei Chen.)}}
\thanks{$^{1}$Zhuo Li, Junjia Liu, Dianxi Li, Tao Teng and Fei Chen are with the Department of Mechanical and Automation Engineering, T-Stone Robotics Institute, The Chinese University of Hong Kong, Hong Kong (e-mail: zli@mae.cuhk.edu.hk; jjliu@mae.cuhk.edu.hk; dianxili@cuhk.edu.hk; tao.teng@ieee.org; f.chen@ieee.org).}
\thanks{$^{2}$Tao Teng and Fei Chen are also with the Hong Kong Center for Logistics Robotics, Hong Kong SAR.}
\thanks{$^{3}$Miao Li is with the Department of Technological Sciences, Wuhan University, Hubei, China (e-mail: miao.li@whu.edu.cn).}
\thanks{$^{4}$Sylvain Calinon is with Idiap Research Institute, Martigny, Switzerland (e-mail: sylvain.calinon@idiap.ch).}
\thanks{$^{5}$Darwin Caldwell is with the Department of Advanced Robotics, Istituto Italiano di Tecnologia, Genoa, Italy (e-mail: darwin.caldwell@iit.it).}
}

\begin{document}
\maketitle

\begin{abstract}
Recent work has demonstrated the potential of diffusion models in robot bimanual skill learning. However, existing methods ignore the learning of posture-dependent task features, which are crucial for adapting dual-arm configurations to meet specific force and velocity requirements in dexterous bimanual manipulation. To address this limitation, we propose Manipulability-Aware Diffusion Policy (ManiDP), a novel imitation learning method that not only generates plausible bimanual trajectories, but also optimizes dual-arm configurations to better satisfy posture-dependent task requirements. ManiDP achieves this by extracting bimanual manipulability from expert demonstrations and encoding the encapsulated posture features using Riemannian-based probabilistic models. These encoded posture features are then incorporated into a conditional diffusion process to guide the generation of task-compatible bimanual motion sequences. We evaluate ManiDP on six real-world bimanual tasks, where the experimental results demonstrate a 39.33$\%$ increase in average manipulation success rate and a 0.45 improvement in task compatibility compared to baseline methods. This work highlights the importance of integrating posture-relevant robotic priors into bimanual skill diffusion to enable human-like adaptability and dexterity.

\end{abstract}

\section{INTRODUCTION}
\begin{figure}[t] 
\centering 
\includegraphics[width=0.47\textwidth]{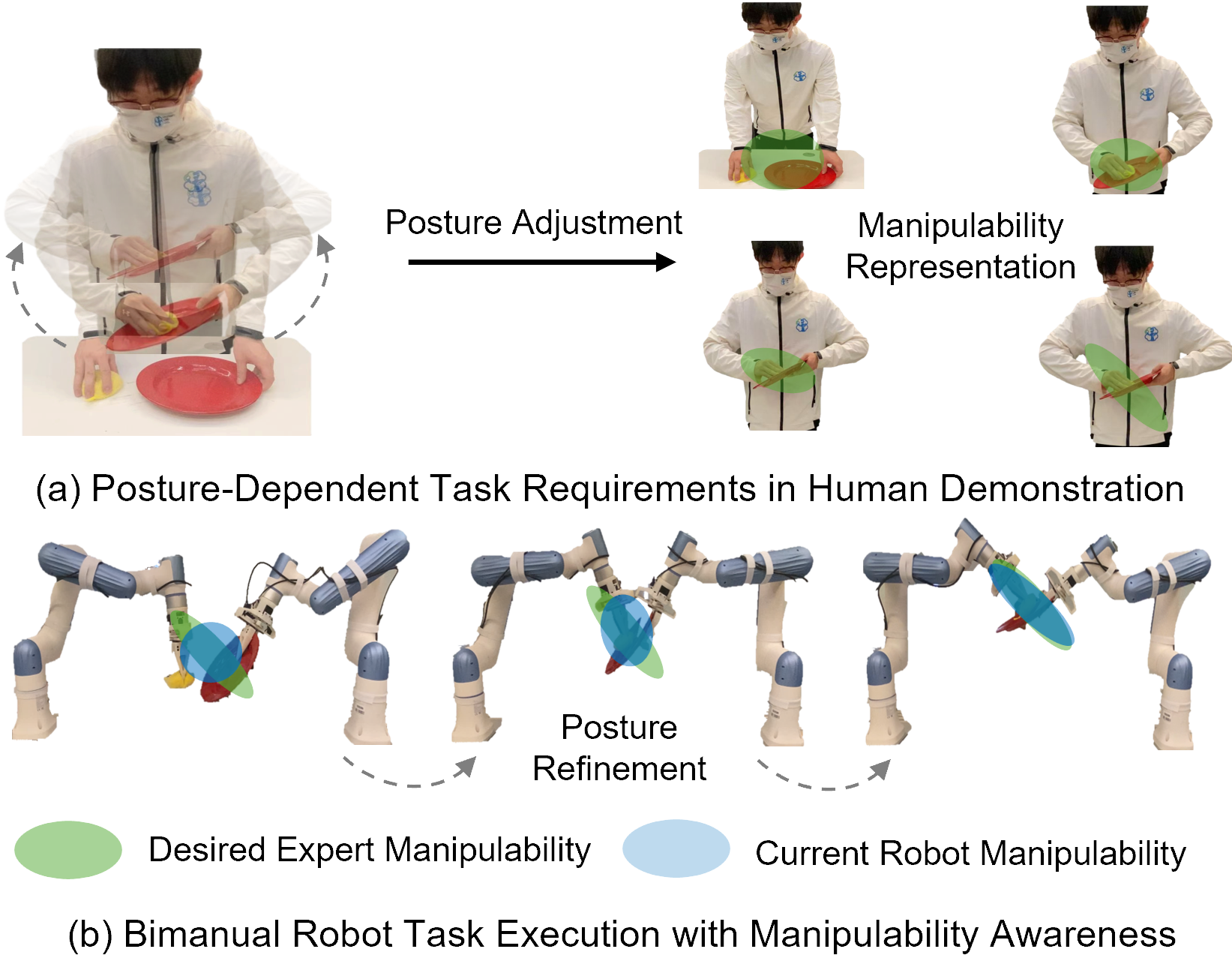} 
\caption{\textbf{Illustration of posture-dependent bimanual task requirements.} (a) Humans naturally adjust their arm posture to optimize manipulation performance. Manipulability serves as a key metric to represent this task-related posture adaptation. (b) Robots should therefore have manipulability awareness to satisfy posture-dependent task requirements.} 
\label{Fig1} 
\end{figure}

Arm posture significantly influences human and robot performance in bimanual manipulation, as appropriate configurations enable efficient velocity and force exertion along task-relevant directions. For instance, when wiping a plate, humans naturally raise their elbows to apply greater lateral force, ensuring effective surface cleaning (see Figure \ref{Fig1}). This adaptive posture adjustment, guided by cognitive task understanding, optimally aligns arm motion characteristics with task demands. In robotics, manipulability \cite{yoshikawa1985manipulability} serves as a geometric descriptor that quantifies the robot's dexterity to generate motion in different directions. It provides a measure of how well a robot's posture aligns with task demands and can thus be used to characterize posture variations during manipulation. Although previous research has explored the integration of manipulability into skill learning to address posture-related task demands \cite{abu2020geometry,jaquier2021geometry,sun2023framework,liu2022robot,li2023planning}, these approaches often rely on traditional parametric models, such as DMP \cite{abu2020geometry}, which lack the expressiveness needed to handle the high-dimensional action space and the multimodal dual-arm coordination inherent in bimanual tasks.

Diffusion models have emerged as powerful tools for bimanual imitation learning, demonstrating strong capabilities in handling complex demonstrations \cite{chi2023diffusion,dastider2024apex,lin2024learning}. However, existing approaches primarily focus on imitating low-level trajectory patterns from expert demonstrations while overlooking the high-level posture features inherent in expert task cognition. This limitation reduces their effectiveness in posture-dependent manipulations, where task success critically depends on achieving appropriate arm configurations to meet specific directional force and velocity requirements.

To address this limitation, we propose ManiDP, a novel manipulability-aware diffusion policy for posture-dependent bimanual manipulation. ManiDP generates dual-arm actions that are both
plausible and task-compatible by integrating manipulability learning into bimanual trajectory diffusion. Specifically, we first analyze appropriate representations for characterizing Bimanual Manipulability Ellipsoids (BMEs) under different dual-arm coordination. For symmetric tasks, we introduce bimanual absolute manipulability, while for asymmetric tasks, we define bimanual relative manipulability, which is derived from relative Jacobians \cite{jamisola2015more}. Building on these representations, ManiDP encodes expert BMEs using Riemannian-based probabilistic models on the Symmetric Positive Definite (SPD) manifold, which account for the geometric structure of BMEs to enable efficient learning and reproduction. Finally, ManiDP employs a manipulability-guided diffusion sampling strategy to optimize and refine generated bimanual trajectories with the learned BMEs features. Compared to existing methods, ManiDP enhances bimanual skill learning by not only imitating expert trajectories but also transferring posture-dependent task cognition encapsulated in expert BMEs. This enables more effective and adaptive bimanual manipulation. The key contributions of this work are summarized as follows:
\begin{itemize}
    \item {We propose ManiDP, a novel diffusion policy that incorporates manipulability learning into bimanual trajectory diffusion, enabling the generation of both plausible and task-compatible dual-arm actions.}
    \item {We analyze bimanual manipulability representations across different dual-arm coordination.}
    \item {We conduct a comprehensive evaluation of ManiDP across six real-world bimanual tasks, demonstrating a 39.33$\%$ increase in average manipulation success rate and a 0.45 improvement in task compatibility compared to baseline methods.}
\end{itemize}

\section{RELATED WORK}
\subsection{Manipulability Learning }
Manipulability quantifies a robot arm's dexterity in achieving desired postures and can be learned using various manifold-based methods \cite{abu2020geometry, jaquier2021geometry,sun2023framework, reithmeir2022human}. Fares et al. \cite{abu2020geometry} integrated SPD-based manipulability learning with DMP, enabling adaptation of learned manipulability profiles on different nonlinear velocity scales. Sun et al. \cite{sun2023framework} introduced a Type-2 fuzzy model-based framework for manipulability learning, achieving high accuracy and computational efficiency. Anna et al. \cite{reithmeir2022human} developed a manifold-aware Iterative Closest Point algorithm to align manipulability ellipsoids as points on the Riemannian manifold. However, existing methods focus only on single-arm tasks, overlooking the complexities of bimanual manipulation, where coordinated posture adaptation is critical for task execution. In this work, we systematically explore manipulability representation for bimanual tasks under different dual-arm coordination, and leverage SPD-based probabilistic models to efficiently learn and reproduce BMEs, which bridges the gap in manipulability learning for bimanual manipulation.

\subsection{Diffusion Model for Robotic Manipulation}
Diffusion models have advantages in handling high-dimensional and multimodal distributions, leading to application in various robotic manipulation tasks, such as dexterous grasp synthesis \cite{barad2024graspldm,weng2024dexdiffuser, li2025language}, motion planning \cite{urain2023se,carvalhomotion}, and bimanual manipulation \cite{chi2023diffusion,dastider2024apex,lin2024learning}. Chi et al. \cite{chi2023diffusion} reformulates visuomotor policy learning as a conditional denoising diffusion process, enabling robots to generate bimanual actions. Dastider et al. \cite{dastider2024apex} introduce a collision-free latent diffusion model tailored for ambidextrous manipulation tasks, using classifier-guidance techniques to integrate obstacle information and ensure safety and feasibility in generated trajectories. Toru et al. \cite{lin2024learning} focus on learning visuotactile skills for multifingered bimanual manipulation by leveraging low-cost teleoperation systems. ManiDP distinguishes itself from previous methods by integrating posture-related manipulability awareness into the bimanual trajectory diffusion process, enabling the generation of high-dimensional dual-arm motions that are both feasible and task-compatible.

\section{METHODS}\label{METHODS}
\begin{figure*}[t] 
\centering 
\includegraphics[width=0.95\textwidth]{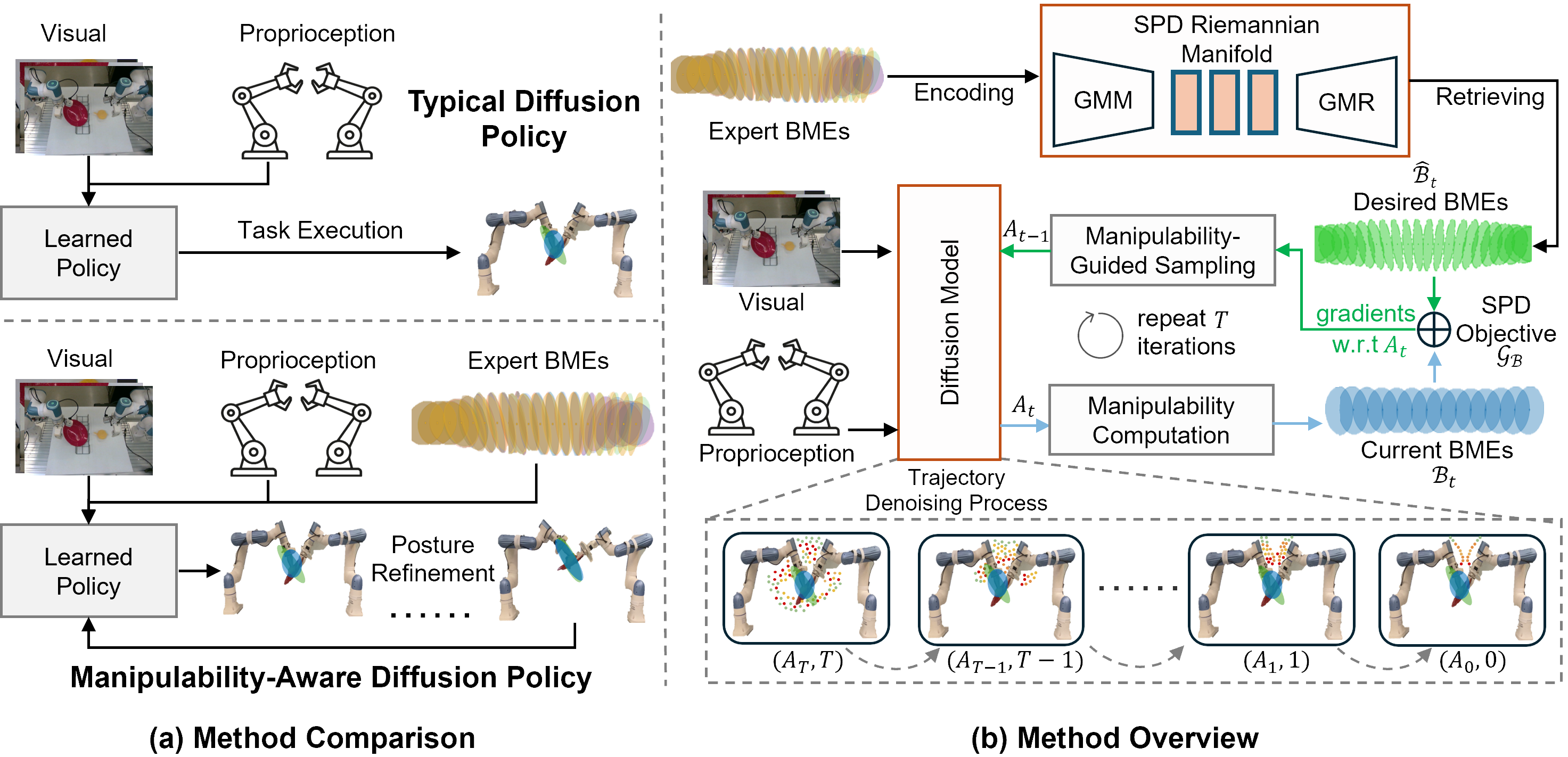}
\caption{\textbf{Overview of the Manipulability-Aware Diffusion Policy (ManiDP).} (a) Unlike the typical diffusion policy that focuses solely on trajectory imitation, ManiDP incorporates BMEs learning into the skill diffusion process to refine dual-arm configuration, enabling enhanced compatibility with posture-dependent task requirements. (b) ManiDP learns a joint distribution of BMEs features and robot trajectories through two key steps: 1) bimanual manipulability learning, where expert BMEs are learned and reproduced using GMM and GMR on the SPD manifold, and 2) manipulability-guided diffusion sampling, where ManiDP iteratively refines the generated bimanual trajectories by minimizing the SPD objective between the current BMEs and the reproduced expert BMEs.}
\label{Fig2} 
\end{figure*}

An overview of ManiDP is shown in Figure \ref{Fig2}. This section begins with a formal problem formulation, followed by a detailed explanation of our proposed method.

\subsection{Problem Formulation} \label{section:gd}
We study bimanual policy learning within the behavior cloning paradigm, where the policy learns from a dataset $D=\{d_i\}_{i=1}^N$, consisting of $N$ samples of bimanual demonstration $d_i$. Each demonstration is an observation-action pair $(\mathcal{O}, \mathcal{A})$, where $\mathcal{O}=\{O_{t-(m-1)}, \dots, O_{t-1}, O_t\}$ captures the past observations over $m$ time steps, and $\mathcal{A} = \{A_t, A_{t+1}, \dots, A_{t+(n-1)}\}$ contains the future actions over $n$ time steps. The observation $O_t = (I_t, S_t, \mathcal{B}_t)$ is composed of visual information $I_t$, robot proprioception state $S_t$, and bimanual manipulability $\mathcal{B}_t$. The robot proprioception state $S_t$ includes the dual-arm joint angle $q_t \in \mathbb{R}^{12}$ and gripper open-close states $g_t \in \mathbb{R}^2$. The bimanual manipulability $\mathcal{B}_t$ is represented as a $3\times3$ SPD matrix, which characterizes the translational manipulability ellipsoid of both arms. The action $A_t \in \mathbb{R}^{14}$ is composed of the joint angles $q_t$ and the gripper states $g_t$. We aim to learn a policy $\pi$ capable of generating manipulability-aware actions $A_t$ based on the current observation $O_t$, enabling task-compatible bimanual manipulation.

\subsection{Bimanual Manipulability Representation}
Bimanual manipulation demands a high level of coordination between both manipulators, which can be broadly classified into symmetric \cite{amadio2019exploiting} and asymmetric tasks \cite{ureche2018constraints}. This section formally defines bimanual manipulability representations tailored to each category. In symmetric tasks, the dual-arm system operates as a unified entity. Thus, the manipulability of this combined system extends the concept from single-arm systems \cite{yoshikawa1985manipulability}. Following the formulation in \cite{ren2024enabling}, we employ Bimanual Absolute Manipulability (BAM) to characterize posture variations in symmetric tasks:

\begin{equation}
\label{equ_1}
\setlength{\jot}{10pt}
\begin{aligned}
v_a^T \left( (G^\dagger)^T J_e W_{v,a}^T W_{v,a} J_e^T G^\dagger \right)^{-1} v_a &\leq 1, \\
h_a^T \left( (G^\dagger)^T J_e W_{h,a}^T W_{h,a} J_e^T G^\dagger \right) h_a &\leq 1,
\end{aligned}
\end{equation}
where $v_a$ and $h_a$ represent the absolute velocity and the external wrench in the world frame.
$J_e=\text{diag}(J_l, J_r) \in \mathbb{R}^{2m \times 2n}$  is the extended Jacobian matrix of the dual arms. $G = \begin{bmatrix} G_l, G_r \end{bmatrix} \in \mathbb{R}^{m \times 2m}$ is the grasp matrix. $m$ and $n$ denote the dimensions of the task space and the robot's joint space, respectively. $W_{k, a}$ for $k=(v, h)$ is the corresponding weighting matrix that addresses nonhomogeneity in spatial translational and rotational manipulability.

However, the BAM representation is inadequate for asymmetric tasks, which requires a more refined representation to capture relative motion and dynamic coordination between the arms. To this end, we introduce Bimanual Relative Manipulability (BRM), which extends the concept of the relative Jacobian \cite{jamisola2015more}. BRM provides a precise characterization of the relative motion between the two arms and is defined as:

\begin{equation}
\setlength{\jot}{10pt}
\label{equ_2}
\begin{aligned}
v_a^T \left(J_{rel} W_{v,a}^T W_{v,a} J_{rel}^T \right)^{-1} v_a &\leq 1, \\
h_a^T \left( J_{rel} W_{h,a}^T W_{h,a} J_{rel}^T \right) h_a &\leq 1,
\end{aligned}
\end{equation}
where $J_{rel}$ represents the relative Jacobian of the dual-arm system, defined as:

\begin{equation}
J_{rel} = \begin{bmatrix}
-{}^{E_{l}}\Psi_{E_{r}} {}^{E_{l}}\Omega_{B_{l}} J_l & {}^{E_{l}}\Omega_{B_{r}} J_r
\end{bmatrix},
\end{equation}
where
\begin{equation}
{}^i\Psi_j = 
\begin{bmatrix}
I & -S({}^ip_j) \\
0 & I
\end{bmatrix}
\quad \text{and} \quad
{}^i\Omega_j = 
\begin{bmatrix}
{}^iR_j & 0 \\
0 & {}^iR_j
\end{bmatrix}.
\end{equation}

Note that ${}^i\Psi_j$is the wrench transformation matrix between the left and right end-effector frames (i.e., $E_l$ and $E_r$), 
${}^i\Omega_j$ is the rotation transformation matrix describing the relationship between the end-effector frames and the base frames (i.e., $B_l$ and $B_r$). $S({}^ip_j)$ is the skew-symmetric matrix of the position vector ${}^ip_j$. The BRM representation captures the fine-grained coordination between the two arms via relative Jacobian $J_{rel}$, making it suitable for asymmetrical tasks where independent yet synchronized arm movements are critical. The BAM and BRM representations offer complementary perspectives on bimanual manipulability. While BAM characterizes the global manipulability of the dual-arm system in symmetric tasks, BRM captures the relative coordination between the arms in asymmetric tasks. These representations form the foundation for learning bimanual posture features, as discussed in the following sections. 

\subsection{Bimanual Manipulability Learning} 
\begin{figure}[!t]
\centering
\includegraphics[width=0.9\linewidth]{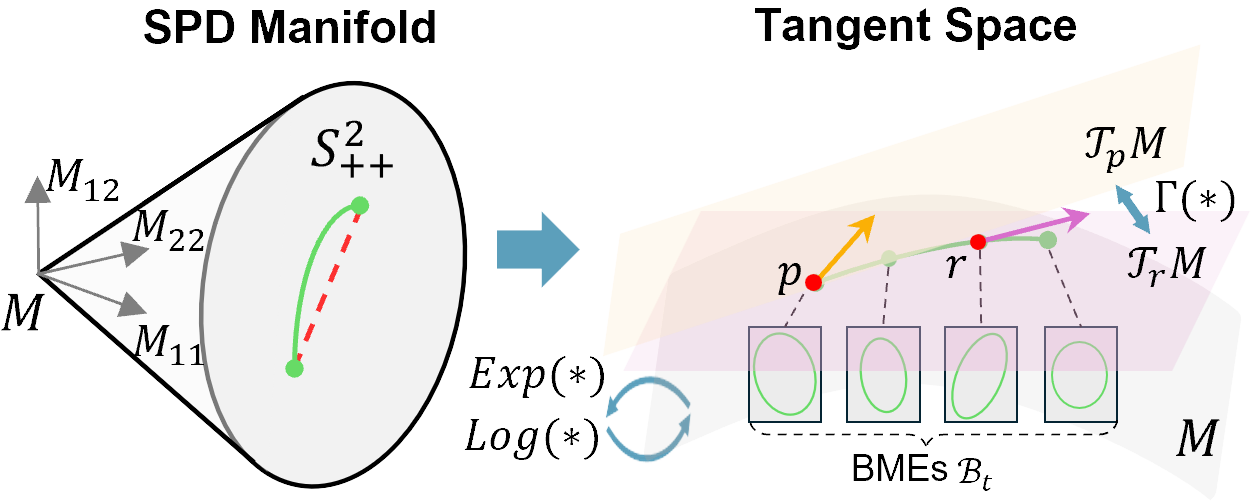}
\caption{\textbf{Visualization of BMEs on the SPD manifold $\mathcal{S}_{++}^2$.} BMEs follow a geodesic path (green curve) on the cone-shaped $\mathcal{S}_{++}^2$, contrasting with the Euclidean path (red dashed line). Each geodesic point represents a bimanual manipulability matrix with an associated tangent space. SPD-GMM and SPD-GMR learn BMEs using three Riemannian operations: logarithmic map $\text{Log}(*)$, exponential map $\text{Exp}(*)$ and parallel transport $\Gamma(*)$.}
\label{fig3}
\end{figure}

Given an appropriate representation of expert BMEs, the key challenge lies in accurately encoding and retrieving BMEs features that capture posture-dependent task requirements. Since BMEs are represented as a set of SPD matrices, forming a geodesic within the SPD manifold (see Figure \ref{fig3}), their intrinsic geometric structure imposes constraints that necessitate a specialized learning model. To this end, we employ an SPD-aware Gaussian Mixture Model combined (SPD-GMM) with Gaussian Mixture Regression (SPD-GMR) \cite{jaquier2021geometry}, which explicitly accounts for the SPD geometry of the BMEs.

To model BMEs in the SPD space, the conventional GMM from the Euclidean space is first extended to the SPD manifold as follows:
\begin{equation}
p(\mathcal{B}) = \sum_{k=1}^{K} \pi_k \mathcal{N}_{M}(\mathcal{B}|\boldsymbol{\Xi}_k, \boldsymbol{\mathcal{S}}_k),
\end{equation}
with $K$ being the number of components of the model, and $\pi_k$ representing the priors such that $\sum_{k=1}^{K} \pi_k = 1$. $\boldsymbol{\Xi}_k$ and $\boldsymbol{\mathcal{S}}_k$ represents the model mean and covriance tensor. $\mathcal{N}_{M}$ represents a multivariate distribution on the SPD manifold $M$, which can be expressed as follows:
\begin{multline}
\mathcal{N}_{M}(\mathcal{B}|\boldsymbol{\Xi}, \boldsymbol{\mathcal{S}}) = \\ 
\frac{1}{\sqrt{(2\pi)^{\tilde{D}} |\boldsymbol{\mathcal{S}}|}}
\exp\left( -\frac{1}{2} \operatorname{Log}_{\boldsymbol{\Xi}}(\mathcal{B}) \mathbf{S}^{-1} 
\operatorname{Log}_{\boldsymbol{\Xi}}(\mathcal{B}) \right).
\end{multline}

The parameters of the SPD-GMM are estimated using the Expectation-Maximization (EM) algorithm \cite{jaquier2021geometry}. In the E-step, the posterior probabilities of each model component are computed, while the M-step updates the parameters $\boldsymbol{\Xi}_k$, $S_k$ and $\pi_k$ on the SPD manifold. We then use time-driven SPD-GMR to conditionally generate the desired BMEs profile. SPD-GMR approximates the conditional distribution of BMEs using the learned SPD Gaussian component:

\begin{equation}
p(\mathcal{B}|t) \sim \mathcal{N}_{M}(\hat{\boldsymbol{\Xi}}_{\mathcal{O}\mathcal{O}}, \hat{\boldsymbol{\mathcal{S}}}_{\mathcal{O}\mathcal{O}}^{\mathcal{O}\mathcal{O}}),
\end{equation}
where $\hat{\boldsymbol{\Xi}}_{\mathcal{O}\mathcal{O}}$ and $\hat{\boldsymbol{\mathcal{S}}}_{\mathcal{O}\mathcal{O}}^{\mathcal{O}\mathcal{O}}$ are the mean and covariance tensor of the conditional probability distribution. By leveraging SPD-GMR, we can reproduce the desired BMEs based on the learned SPD-GMM model, effectively capturing the posture-dependent task characteristics.

\subsection{Manipulability-Guided Diffusion Sampling}
\begin{algorithm}[htbp]
\label{algorithm1}
\DontPrintSemicolon
\caption{Manipulability-Guided Bimanual Trajectory Sampling}
\texttt{// diffusion model training}\\
\textbf{Input:} Bimanual Demonstrations $D = \{d_i\}_{i=1}^N$ \\
\Repeat{converged;}{
    Sample a demonstration $d = (\mathcal{O}, \mathcal{A})$ from $D$ \; 
    $A_0 \sim p_\theta(A_0{\mid}\mathcal{O})$ \;
    $\epsilon \sim \mathcal{N}(0, I), \; t \sim \mathcal{U}(\{1, \dots, T\})$ \;
    $A_t = \sqrt{\hat{\alpha}_t} A_0 + \sqrt{1 - \hat{\alpha}_t} \epsilon$ \;
    $\theta = \theta - \eta \nabla_\theta \| \epsilon - \epsilon_\theta(A_t, t, \mathcal{O}) \|_2^2$ \;
}
\texttt{// manipulability-guiede sampling}\\
\textbf{Input:} Trained Diffusion Model $p_\theta(A_0{\mid}\mathcal{O})$,     Manipulability Objective $\mathcal{G}_\mathcal{B}$ and Gradient Scale $\lambda$

\For{$t = T, \ldots, 1$}{
    $\mu = \mu_\theta(A_t, t, \mathcal{O}), \; \Sigma = \Sigma_\theta(A_t, t, \mathcal{O})$ \;
    $\mathbf{g} = \nabla_{A_t} \log p_\phi(\mathcal{G}_\mathcal{B}{\mid}A_t, \mathcal{O}) \big|_{A_t = \mu}$ \;
    $A_{t-1} = \mathcal{N}(A_{t-1}; \mu + \lambda \Sigma \mathbf{g}, \Sigma)$ 
}
\textbf{end for}

\Return $A_0$
\end{algorithm}
The core idea of ManiDP is to leverage the reproduced expert BMEs to guide the diffusion-based bimanual trajectory sampling process, which can be formalized as follows:
\begin{equation}
    \begin{aligned}
    p(A_0{\mid}\mathcal{O}, \mathcal{G}_\mathcal{B}) &= 
    \frac{p_\theta(A_0{\mid}\mathcal{O})p_\phi(\mathcal{G}_\mathcal{B}{\mid}A_0, \mathcal{O})}{p(\mathcal{G}_\mathcal{B}{\mid}\mathcal{O})} \\
    &\propto p_\theta(A_0{\mid}\mathcal{O})p_\phi(\mathcal{G}_\mathcal{B}{\mid}A_0, \mathcal{O}).
    \end{aligned}
\end{equation}

Here, $p_\theta(A_0{\mid}\mathcal{O})$ characterizes the probability distribution for generating bimanual actions with the observation condition $\mathcal{O}$. It can be learned using a conditional diffusion model \cite{chi2023diffusion} from expert demonstrations with an iterative denoising process:
\begin{equation}
\begin{aligned}
    p_\theta(A_0{\mid}\mathcal{O}) &= p(A_T) \prod_{t=1}^T p(A_{t-1}{\mid}A_t, \mathcal{O}), \\
    p(A_{t-1}{\mid}A_t, \mathcal{O}) &= \mathcal{N}(A_{t-1}; \mu_\theta(A_t, t, \mathcal{O}), \Sigma_\theta(A_t, t, \mathcal{O})).
\end{aligned}
\end{equation}

Meanwhile, $p_\phi(\mathcal{G}_\mathcal{B}{\mid}A_0, \mathcal{O})$ represents the probability distribution of reaching the desired BMEs with the sampled bimanual actions, where $\mathcal{G}_\mathcal{B}$ is the manipulability objective function. Considering the manifold properties of BMEs, $\mathcal{G}_\mathcal{B}$ is designed based on the SPD logarithmic map distance:
\begin{equation}
\label{equ_10}
    \mathcal{G}_\mathcal{B} = \| \text{Log}_{\mathcal{B}_{t}}(\hat{\mathcal{B}_t}) \|_F^2,
\end{equation}
where $\hat{\mathcal{B}_t}$ is the reproduced expert BMEs from SPD-GMR, and $\mathcal{B}_t$ is the calculated BMEs of current bimanual actions. $\mathcal{G}_\mathcal{B}$ ensures a precise alignment by measuring the distance between $\mathcal{B}_t$ and $\hat{\mathcal{B}_t}$ on the SPD manifold. With the learned $ p_\theta(A_0{\mid}\mathcal{O})$, we then sample the target probability distribution $p(A_0{\mid}\mathcal{O}, \mathcal{G}_\mathcal{B})$ taking advantage of the flexible conditioning of the diffusion model \cite{dhariwal2021diffusion}. Specifically, we approximate $p_\phi(\mathcal{G}_\mathcal{B}{\mid}A_t, \mathcal{O})$ using the Taylor expansion around $A_t = \mu$ at timestep $t$ as:
\begin{equation}
    \log p_\phi(\mathcal{G}_\mathcal{B}{\mid}A_t, \mathcal{O}) \approx (A_t - \mu) \mathbf{g} + C,
\end{equation}
where $C$ is a constant, $\mu = \mu_\theta(A_t, t, \mathcal{O})$ and $\Sigma = \Sigma_\theta(A_t, t, \mathcal{O})$ are the inferred parameters of the original diffusion process, and:
\begin{equation}
  \mathbf{g} = \nabla_{A_t} \log p_\phi(\mathcal{G}_\mathcal{B}{\mid}A_t, \mathcal{O}) \big|_{A_t = \mu} .
\end{equation}

Therefore, we have：
\begin{equation}
\label{equ_13}
    p(A_{t-1}{\mid}A_t, \mathcal{O}, \mathcal{G}_\mathcal{B}) = \mathcal{N}(A_{t-1}; \mu + \lambda \Sigma \mathbf{g}, \Sigma),
\end{equation}
where $\lambda$ is the scaling factor for the guidance. With Equation \ref{equ_13}, we can sample $A_t$ with the guidance of expert BMEs and progressively optimize the generated action to meet posture-dependent requirements. The detailed diffusion sampling process is summarized in Algorithm 1.

\section{EXPERIMENTS}
We conduct comprehensive experiments to investigate the following questions: 1) Can ManiDP accurately learn and reproduce BMEs under different dual-arm coordination? 2) Can ManiDP satisfy posture-dependent task requirements while generating high-dimensional bimanual trajectories? 3) How does the performance of ManiDP compare with that of the baseline methods?

\subsection{Setup}
The experiments were carried out using the dual arm system DOBOT X-Trainer, which comprises two 6-DoF Nova2 robotic arms equipped with 1-DoF grippers. The system utilized three Intel RealSense D405 cameras to capture RGB image observations. Six bimanual tasks were selected for the experiment (see Figure \ref{Fig6}), categorized by: 1) symmetric tasks, which required synchronized coordination between both arms; and 2) asymmetric tasks, where the two arms performed different functional roles in coordination. Task demonstrations were collected using the X-Trainer master-slave teleoperation system and recorded at 10 Hz.


\textit{1) Experiment Procedure:} We conducted two series of experiments to evaluate the proposed method. First, we assessed ManiDP's ability to capture BMEs characteristics through bimanual manipulability learning experiments. Two representative tasks with distinct dual-arm coordination were selected: tower hanging (symmetric) and plate wiping (asymmetric). The BAM representation (Equation \ref{equ_1}) was used for the symmetric tower hanging task, while the BRM representation (Equation \ref{equ_2}) was applied to the asymmetric plate wiping task. Each task was demonstrated five times for learning and reproduction of the corresponding BMEs. Following this, we evaluated ManiDP's overall performance in the real-robot experiments across all six tasks.


\begin{table}[!t]
\caption{Manipulability reproduction accuracy of BMEs learning experiments. ManiDP achieves the highest MRA in both symmetric and asymmetric tasks.}
\label{tab1}
\centering
\resizebox{\columnwidth}{!}{%
\begin{tabular*}{\columnwidth}{@{\extracolsep{1.5mm}} @{}ccccc@{}}
\toprule
\multirow{4}{*}{\centering \makecell{\textbf{Metric}}} & \multicolumn{2}{c}{\textbf{GMM-GMR}}                     & \multicolumn{2}{c}{\textbf{ManiDP (Ours)}}                              \\ \cmidrule(lr){2-3} \cmidrule(lr){4-5}
                              & \makecell{Plate\\Wiping} & \makecell{Tower\\Hanging} & \makecell{Plate\\Wiping} & \makecell{Tower\\Hanging} \\ \midrule
\makecell{Avg.\\MRA($\uparrow$)}                & 0.58 {\scriptsize$\pm$ 0.02}        & 0.65 {\scriptsize$\pm$ 0.03}           & \textbf{0.91 {\scriptsize$\pm$ 0.01}}                 & \textbf{0.93 {\scriptsize$\pm$ 0.04}}                    \\ \bottomrule
\end{tabular*}%
}
\end{table}

 \begin{figure}[!t]
\centering
\includegraphics[width=1\linewidth]{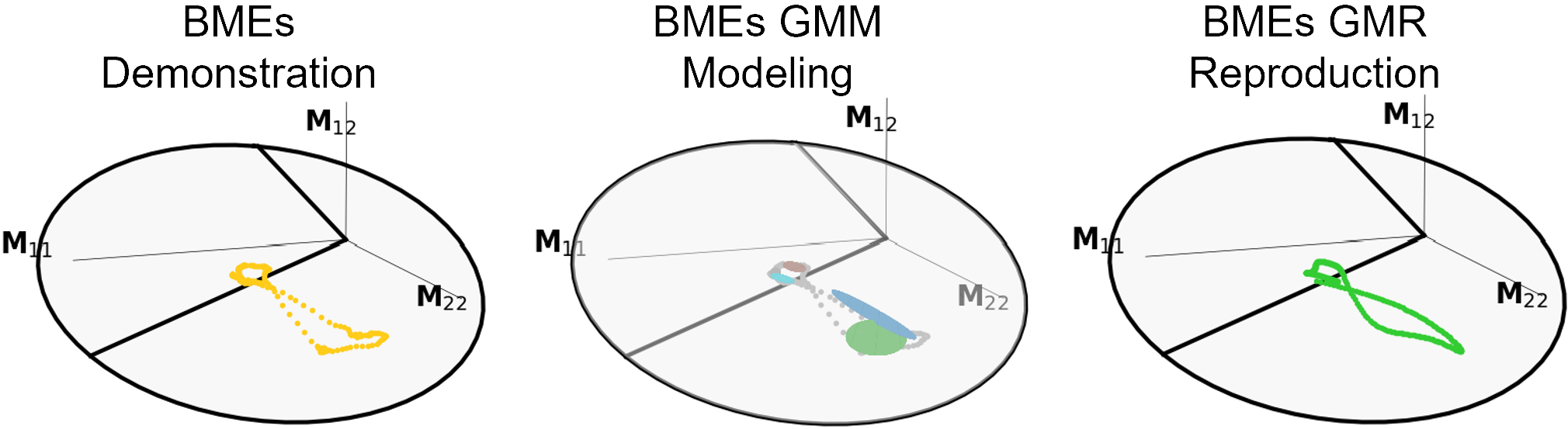}
\caption{\textbf{BMEs learning results of Tower Hanging task on SPD manifold.} ManiDP can generate geometrically plausible BME sequences within the cone-shaped SPD manifold by considering the intrinsic Riemannian properties of BMEs.}
\label{fig4}
\end{figure}

\begin{figure}[!t]
\centering
\includegraphics[width=1\linewidth]{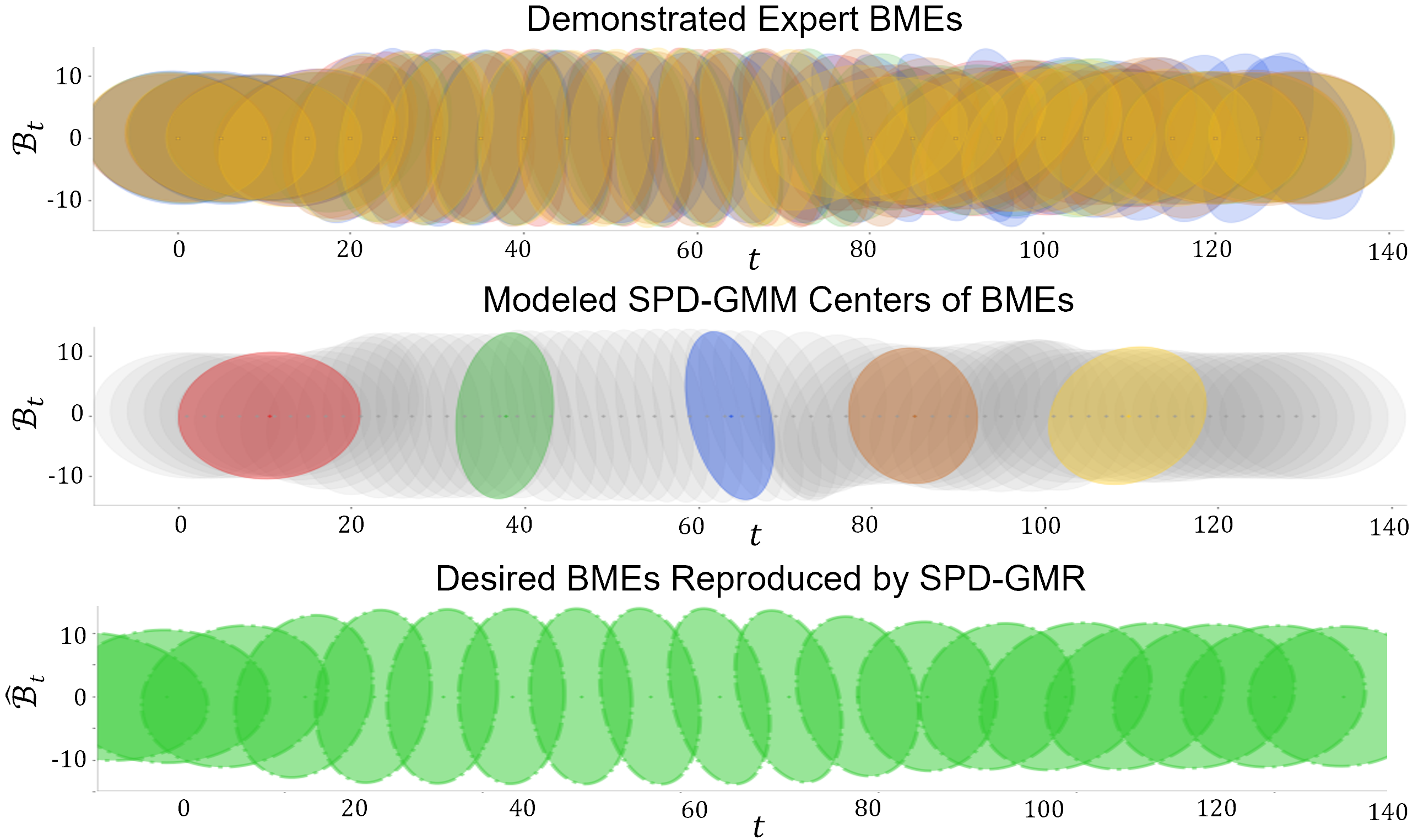}
\caption{\textbf{BMEs learning results of Tower Hanging task on the time domain.} The results show that ManiDP can capture the key states of expert BMEs and then accurately reproduce them at each time step.}
\label{fig5}
\end{figure}

\textit{2) Evaluation Metrics:} Three quantitative metrics are used to evaluate our approach:

\begin{figure*}[t] 
\centering 
\includegraphics[width=1\textwidth]{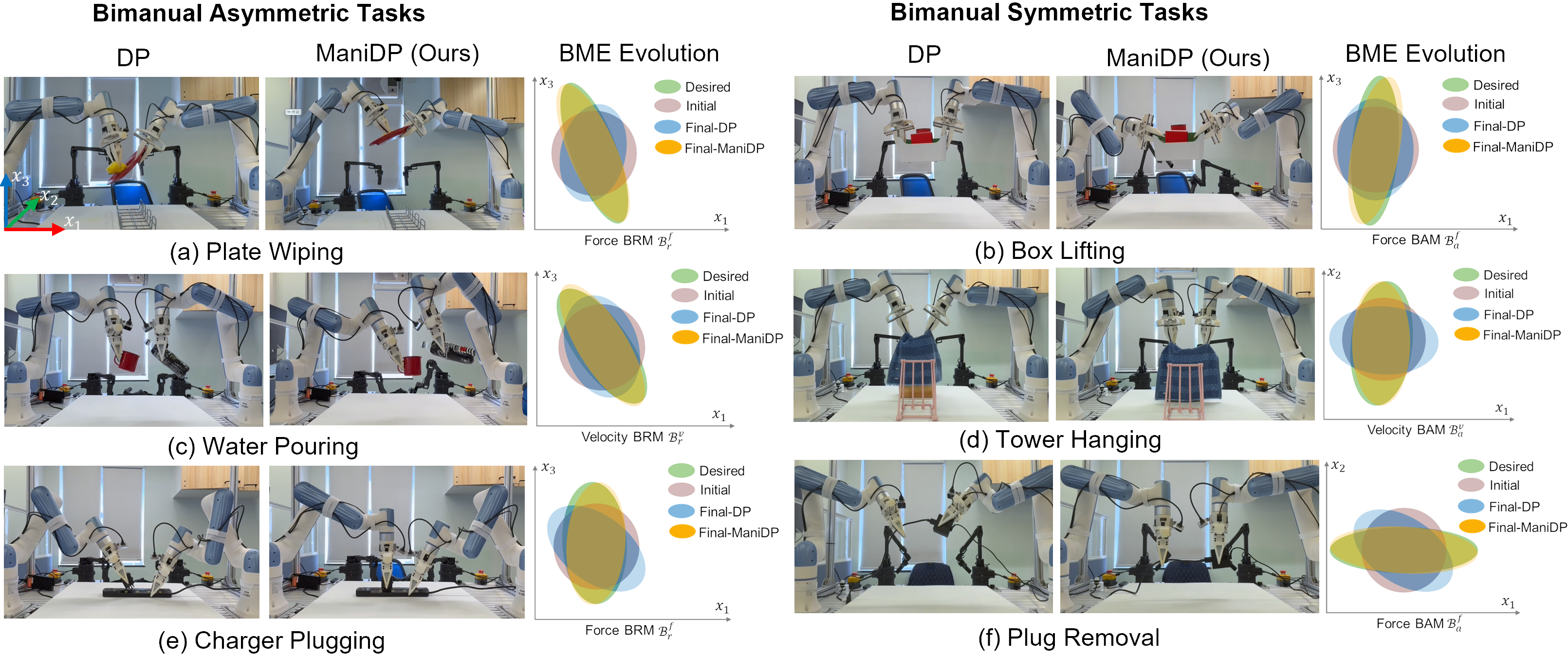}
\caption{\textbf{Qualitative results of real robot experiments.} Compared to the DP baseline, ManiDP can adapt dual-arm postures to minimize the discrepancy between final and desired BMEs. This enhances velocity control and force exertion along task-relevant directions, thereby improving manipulation performance. Note that only the most representative dimension of BMEs is displayed.}
\label{Fig6} 
\end{figure*}

\begin{table*}[!t]
\caption{Quantitative results of real robot experiments. ManiDP outperforms baselines across all tasks in terms of MSR and TCI. The results validate the superiority of our method in generating feasible and task-compatible dual-arm actions.}
\label{tab2}
\centering
\resizebox{\textwidth}{!}{%
\begin{tabular*}{\textwidth}{@{\extracolsep{\fill}} @{}cccccccccc@{}}
\toprule
\multirow{3}{*}{\centering \textbf{\makecell{Tasks \\ (100 demo.)}}} & \multicolumn{2}{c}{\textbf{ACT}}                     & \multicolumn{2}{c}{\textbf{DP}}                              & \multicolumn{2}{c}{\textbf{GMM-GMR}}                         & \multicolumn{2}{c}{\textbf{ManiDP (Ours)}}                         \\ \cmidrule(lr){2-3} \cmidrule(lr){4-5} \cmidrule(lr){6-7} \cmidrule(lr){8-9}
                              & \makecell{Avg.\\MSR($\uparrow$)} & \makecell{Avg.\\TCI($\uparrow$)} & \makecell{Avg.\\MSR($\uparrow$)} & \makecell{Avg.\\TCI($\uparrow$)} & \makecell{Avg.\\MSR($\uparrow$)} & \makecell{Avg.\\TCI($\uparrow$)} & \makecell{Avg.\\MSR($\uparrow$)} & \makecell{Avg.\\TCI($\uparrow$)} \\ \midrule
Plate Wiping                 & 28/50                 & 0.53 \scriptsize$\pm$ 0.02             & 29/50                 & 0.57 \scriptsize$\pm$ 0.05             & 15/50                 & 0.45 \scriptsize$\pm$ 0.03             & 44/50       & 0.93 \scriptsize$\pm$ 0.01    \\
Plug Removal                 & 21/50                 & 0.45 \scriptsize$\pm$ 0.08             & 23/50                 & 0.47 \scriptsize$\pm$ 0.02             & 11/50                 & 0.42 \scriptsize$\pm$ 0.01             & 40/50       & 0.94 \scriptsize$\pm$ 0.02    \\
Water Pouring                & 38/50                 & 0.62 \scriptsize$\pm$ 0.03             & 39/50                 & 0.63 \scriptsize$\pm$ 0.06             & 20/50                 & 0.48 \scriptsize$\pm$ 0.02             & 47/50        & 0.95 \scriptsize$\pm$ 0.01    \\
Box Lifting              & 31/50                 & 0.55 \scriptsize$\pm$ 0.09             & 31/50                 & 0.53 \scriptsize$\pm$ 0.04             & 15/50                 & 0.38 \scriptsize$\pm$ 0.07             & 45/50        & 0.89 \scriptsize$\pm$ 0.05    \\
Towel Hanging                & 25/50                 & 0.50 \scriptsize$\pm$ 0.05             & 27/50                 & 0.52 \scriptsize$\pm$ 0.01             & 13/50                & 0.39 \scriptsize$\pm$ 0.03             & 41/50       & 0.91 \scriptsize$\pm$ 0.04  \\
Charger Plugging             & 17/50                 & 0.45 \scriptsize$\pm$ 0.03             & 20/50                 & 0.48 \scriptsize$\pm$ 0.06             & 8/50                 & 0.28 \scriptsize$\pm$ 0.01             & 38/50        & 0.87 \scriptsize$\pm$ 0.07    \\ \midrule
\textbf{Overall}             & 53.33$\%$               & 0.51 \scriptsize$\pm$ 0.05             & 56.33$\%$                 & 0.53 \scriptsize$\pm$ 0.04             & 27.33$\%$                 & 0.40 \scriptsize$\pm$ 0.03             & \textbf{85.00$\%$}        & \textbf{0.92 \scriptsize$\pm$ 0.03}    \\ \bottomrule
\end{tabular*}%
}
\end{table*}

\begin{itemize}
    \item \textbf{Manipulability Reproduce Accuracy (MRA):} Measures the similarity between reproduced and demonstrated BMEs using the SPD logarithmic map distance：
    \begin{equation}
        \text{MRA} = e^{-\mathcal{G}_\mathcal{B}}
    \end{equation}
    \item \textbf{Task Compatibility Index (TCI) \cite{chiu1988task}:} Assesses the alignment of dual-arm postures with task requirements. In our experiments, TCI is normalized to [0,1], with higher values indicating better posture alignment for efficient force and velocity transmission.
    \item \textbf{Manipulation Success Rate (MSR):} Computes the percentage of successful task executions using the generated manipulation trajectories.
\end{itemize}

\textit{3) Implementation Details:} The SPD-GMM and SPD-GMR models were configured with five components. For trajectory generation, a UNet-based DDIM \cite{chi2023diffusion} was used as the diffusion model, with a squared cosine scheduler for noise scheduling. The number of diffusion and denoising steps was set to 100 and 10, respectively. The model was trained using the Adam optimizer with a learning rate of 0.0001, weight decay of 0.00001, and a batch size of 128. Following \cite{chi2023diffusion}, an exponential moving average of the model weights was maintained to enhance stability during training and ensure reliable trajectory generation for diverse tasks.

\subsection{Results}
\textit{1) Bimanual Manipulability Learning:} We compare ManiDP with the classical Euclidean GMM-GMR. As shown in Table \ref{tab1}, ManiDP achieves the highest MRA in both tasks, demonstrating its effectiveness in learning and reproducing BMEs under varying dual-arm coordination. Notably, it outperforms the baseline by an average MRA margin of 0.31. This can be attributed to its explicit consideration of BME geometric properties. Unlike baseline methods that ignore manifold structure, ManiDP preserves the intrinsic Riemannian characteristics of BMEs, enabling the generation of geometrically plausible BME sequences within the cone-shaped SPD manifold (see Figure \ref{fig4}). An important observation is ManiDP's consistent performance across symmetric and asymmetric tasks, with only a minor MRA difference of 0.02. This suggests that the BRM representation effectively captures relative motion and dynamic coordination between the arms, allowing ManiDP to generalize well across different coordination patterns. Figure \ref{fig5} illustrates that ManiDP effectively captures the key states of the demonstrated BMEs, as represented by the GMM centers, and accurately reproduces expert BMEs at each time step. This temporal consistency highlights ManiDP's ability to maintain high fidelity in transitioning between key manipulability states, ensuring posture-relevant precision throughout task execution.

\textit{2) Real-Robot Experiments:} We compare ManiDP with three baseline methods: (1) DP \cite{chi2023diffusion}, the original Diffusion Policy; (2) ACT \cite{zhao2023learning}, the Action Chunking Transformer trained as a conditional VAE; and (3) Classical GMM-GMR, which learns in Euclidean space. Table \ref{tab2} shows the experimental result for each task with 100 demonstrations over 50 test trials. The results illustrate that ManiDP achieves superior performance compared to the baseline methods. In terms of TCI, ManiDP achieves a significant improvement over ACT and DP by margins of 0.41 and 0.39, respectively, demonstrating its effectiveness to generate task-compatible dual-arm postures. Qualitative results in Figure \ref{Fig6} further demonstrates this advantage, where ManiDP enables the robot to autonomously adjust its bimanual configurations to postures that are more conducive to task execution. This improvement can be attributed to the ManiDP's incorporation of manipulability guidance into the diffusion sampling process, which minimizes the discrepancy between the final and desired BMEs (see Figure \ref{Fig6}). In contrast, the baseline methods exhibit a larger BMEs gap, resulting in inappropriate dual-arm configurations that hinder task execution. For example, in the Plug Removal task, ManiDP optimally aligns the arms to apply effective lateral force for plug detachment, whereas DP fails to generate an appropriate posture, resulting in inefficient force transmission. We find that this inability to generate task-compatible postures is the primary factor contributing to the lower average MSR of the baseline methods (i.e., 53.33$\%$ for ACT and 56.33$\%$ DP). ManiDP achieves higher MSR by leveraging posture-dependent task cognition encapsulated in expert BMEs, leading to consistently superior performance across all tasks.

\subsection{Discussion}
Experimental results demonstrate that ManiDP can accurately learn and reproduce expert BMEs, and then capture the encapsulated posture-dependent features to generate plausible and task-compatible bimanual trajectories. Compared to baseline methods, ManiDP enhances bimanual skill learning by not only imitating expert trajectories but also transferring posture-related task cognition, resulting in more effective and adaptive bimanual manipulation. Furthermore, leveraging the generalizability of manipulability ellipsoids for robot posture representation, ManiDP can adapt to novel robotic platforms without requiring extensive retraining. Overall, ManiDP provides new insights on incorporating posture-relevant robotic priors in trajectory diffusion, advancing bimanual robotic systems toward human-like adaptability and dexterity. It has the potential to extend to a wide range of applications where robotic posture is critical to task success. For example, in industrial assembly tasks, optimizing posture can greatly improve precision and efficiency, particularly when handling delicate or complex components.





\section{CONCLUSIONS}
In this work, we present ManiDP, a novel manipulability-aware diffusion policy for posture-dependent bimanual manipulation. By incorporating manipulability learning into the trajectory diffusion process, ManiDP enables robots to optimize dual-arm configurations to enhancing task compatibility in posture-sensitive scenarios. Extensive real-world experiments demonstrate ManiDP's effectiveness, showcasing its superior performance in both symmetric and asymmetric bimanual tasks compared to state-of-the-art imitation learning baselines. One limitation of our method is its exclusive reliance on velocity and force BMEs for guiding bimanual trajectory diffusion. Future work will investigate the impact of alternative manipulability representations, such as dynamic and impedance BMEs, to further enhance the method's versatility and robustness.

\bibliographystyle{IEEEtran}
\bibliography{references}

\end{document}